\documentclass[conference]{IEEEtran}
\IEEEoverridecommandlockouts

\usepackage{cite}
\usepackage{amsmath,amssymb,amsfonts}
\usepackage{algorithmic}
\usepackage{graphicx}
\usepackage{textcomp}
\usepackage{xcolor}
\usepackage{booktabs}
\usepackage{multirow}
\usepackage{subfigure}
\usepackage{hyperref}
\usepackage{float}

\def\BibTeX{{\rm B\kern-.05em{\sc i\kern-.025em b}\kern-.08em
    T\kern-.1667em\lower.7ex\hbox{E}\kern-.125emX}}

\begin{document}

\title{47B Mixture-of-Experts Beats 671B Dense Models on Chinese Medical Examinations}

\author{
    \IEEEauthorblockN{
        Chiung-Yi Tseng\textsuperscript{1,4*}, 
        Danyang Zhang\textsuperscript{1,5*},
        Tianyang Wang\textsuperscript{1},
        Hongying Luo\textsuperscript{1},
        Lu Chen\textsuperscript{1},\\
        Junming Huang\textsuperscript{1},
        Jibin Guan\textsuperscript{6},
        Junfeng Hao\textsuperscript{6},
        Junhao Song\textsuperscript{7},
        Xinyuan Song\textsuperscript{1},
        Ziqian Bi\textsuperscript{1,2†},
    }
    \\
    \IEEEauthorblockA{\textsuperscript{1}AI Agent Lab, Vokram Group, United Kingdom, ai-agent-lab@vokram.com}
    \IEEEauthorblockA{\textsuperscript{2}Purdue University, United States, bi32@purdue.edu}
    \IEEEauthorblockA{\textsuperscript{4}LuxMuse AI, United States, ctseng@luxmuse.ai}
    \IEEEauthorblockA{\textsuperscript{5}ByteDance Inc, United States, joseph.zhang@bytedance.com}
    \IEEEauthorblockA{\textsuperscript{6}University of Minnesota, United States, jguan@umn.edu, ygzhjf85@gmail.com}
    \IEEEauthorblockA{\textsuperscript{7}Imperial College London, United Kingdom, junhao.song23@imperial.ac.uk}
}
\maketitle

\begin{abstract}
The rapid advancement of large language models (LLMs) has prompted significant interest in their potential applications in medical domains. %
This paper presents a comprehensive benchmark evaluation of 27 state-of-the-art LLMs on Chinese medical examination questions, %
encompassing seven medical specialties across two professional levels. %
We introduce a robust evaluation framework that assesses model performance on 2,800 carefully curated questions from %
cardiovascular, gastroenterology, hematology, infectious diseases, nephrology, neurology, and respiratory medicine domains. %
Our dataset distinguishes between attending physician and senior physician difficulty levels, %
providing nuanced insights into model capabilities across varying complexity. %
Our empirical analysis reveals substantial performance variations among models, %
with Mixtral-8x7B achieving the highest overall accuracy of 74.25\%, followed by DeepSeek-R1-671B at 64.07\%. %
Notably, we observe no consistent correlation between model size and performance, as evidenced by the strong performance %
of smaller mixture-of-experts architectures. The evaluation demonstrates significant performance gaps between %
medical specialties, with models generally performing better on cardiovascular and neurology questions compared to %
gastroenterology and nephrology domains. Furthermore, our analysis indicates minimal performance degradation between %
attending and senior physician levels for top-performing models, suggesting robust generalization capabilities. %
This benchmark provides critical insights for the deployment of LLMs in medical education and clinical decision support systems, %
highlighting both the promise and current limitations of these technologies in specialized medical contexts. %
\end{abstract}

\section{Introduction}

The integration of artificial intelligence (AI) in healthcare has witnessed unprecedented growth, with Large Language Models (LLMs) emerging as particularly promising tools for medical applications \cite{singhal2023large,nori2023capabilities}. These models have demonstrated remarkable capabilities in understanding complex medical literature, answering clinical questions, and assisting in medical education \cite{kung2023performance,gilson2023chatgpt}. However, the evaluation of LLMs in non-English medical contexts, particularly in Chinese healthcare settings, remains underexplored despite China representing one of the world's largest healthcare markets with unique clinical practices and examination standards \cite{zhang2023chinese}.

\begin{figure}[h]
\centering
\includegraphics[width=\columnwidth]{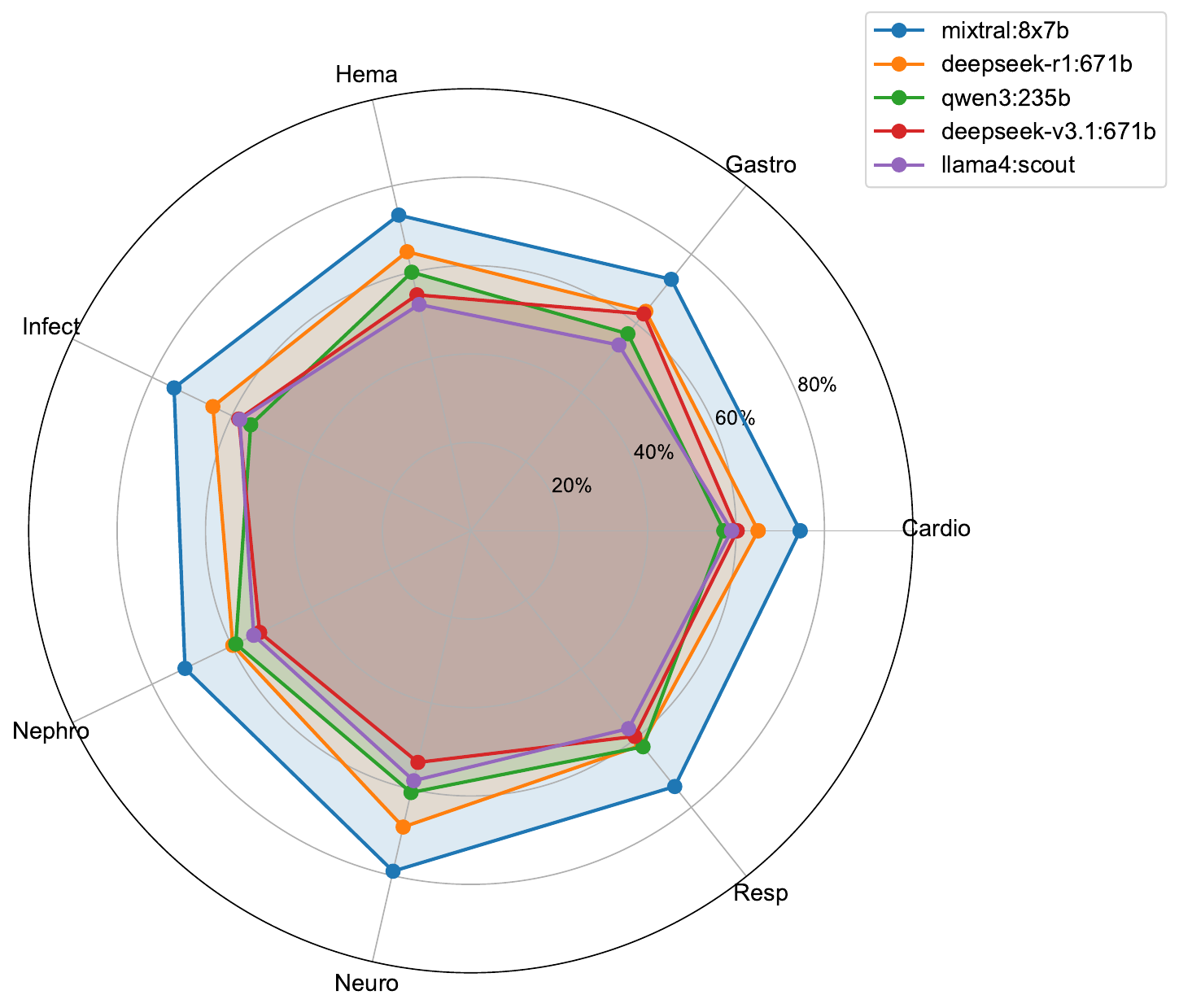}
\caption{Radar chart comparing top 5 models across medical specialties with larger areas indicating better overall performance.}
\label{fig:radar}
\end{figure}

The Chinese medical licensing examination system presents a rigorous framework for assessing medical knowledge across multiple specialties and professional levels \cite{wang2022medical}. This system distinguishes between attending physicians and senior physicians, with each level requiring a comprehensive understanding of specialty-specific knowledge and clinical decision-making capabilities. The examination questions span critical medical domains including cardiovascular medicine, gastroenterology, hematology, infectious diseases, nephrology, neurology, and respiratory medicine, each presenting unique challenges in terminology, diagnostic criteria, and treatment protocols specific to Chinese medical practice \cite{liu2021chinese}.

Despite the growing deployment of LLMs in medical applications, several critical gaps persist in current evaluation methodologies. Existing benchmarks predominantly focus on English-language medical content, potentially overlooking linguistic and cultural nuances essential for effective healthcare delivery in non-English speaking populations \cite{chen2023meditron}. Furthermore, most evaluations fail to differentiate between varying levels of medical expertise, treating all questions with equal complexity regardless of their intended professional audience \cite{pal2022medmcqa}. Comprehensive multi-specialty evaluations that reflect real-world clinical diversity remain scarce, with many studies focusing on single domains or limited question sets \cite{jin2021disease}.

This paper addresses these limitations by introducing a comprehensive benchmark for evaluating LLMs on Chinese medical examination questions. We present a carefully curated dataset of 2,800 Chinese medical examination questions spanning seven major medical specialties across two professional levels, providing unprecedented granularity in assessing model capabilities in specialized medical contexts. Through extensive evaluation of 27 state-of-the-art LLMs, including both open-source and proprietary models with varying architectures and parameter counts, we offer insights into the relationship between model design choices and medical question-answering performance. Our detailed empirical analysis reveals performance patterns across specialties, difficulty levels, and model architectures, identifying key factors that influence success in medical question-answering tasks.

Our findings reveal substantial performance variations among models, with the best-performing model achieving 74.25\% accuracy while others struggle to exceed random guessing baselines. Figure \ref{fig:radar} illustrates the performance profiles of the top five models across all medical specialties, revealing distinct strengths and weaknesses in different domains.

Interestingly, we observe that model size alone does not determine performance, as several smaller mixture-of-experts models outperform their larger dense counterparts. These results have important implications for the deployment of LLMs in medical education and clinical support systems, particularly in Chinese healthcare settings where an accurate understanding of specialty-specific knowledge is crucial. The remainder of this paper is organized as follows: Section II reviews related work, Section III describes our methodology, Section IV details experimental results, Section V discusses implications, and Section VI concludes with future directions.
\section{Related Work}

\subsection{Medical Question Answering Benchmarks}

The evaluation of AI systems on medical knowledge has evolved significantly over the past decade. Early benchmarks such as BioASQ \cite{tsatsaronis2015overview} and MEDIQA \cite{ben2019overview} established foundational frameworks for assessing natural language understanding in biomedical contexts. These datasets primarily focused on information retrieval and question-answering tasks derived from scientific literature, laying groundwork for more sophisticated evaluations.

Recent developments have introduced more comprehensive medical examination-based benchmarks. MedQA \cite{jin2021disease} aggregated medical licensing examinations from multiple countries, including the United States Medical Licensing Examination (USMLE), providing standardized evaluation across different healthcare systems. Similarly, MedMCQA \cite{pal2022medmcqa} contributed over 194,000 multiple-choice questions from Indian medical entrance examinations, enabling large-scale assessment of medical knowledge. PubMedQA \cite{jin2019pubmedqa} focused on research question answering using PubMed abstracts, bridging clinical practice and medical literature.

However, these benchmarks predominantly feature English-language content and Western medical practices. The few Chinese medical benchmarks available, such as CMedQA \cite{zhang2017chinese} and WebMedQA \cite{he2019applying}, primarily focus on consumer health questions rather than professional medical examinations. CBLUE \cite{zhang2022cblue} introduced a comprehensive Chinese biomedical language understanding evaluation, yet lacks the specialty-specific granularity and professional level differentiation critical for assessing clinical expertise.

More recently, medical benchmark construction has shifted toward higher-fidelity, exam-oriented, and multilingual evaluations. Several studies emphasized the need for native-language professional medical benchmarks that avoid translation artifacts and reflect real clinical licensing standards. Large-scale extensions of examination-based datasets have been proposed to better capture specialty-level variance, cognitive difficulty stratification, and clinical reasoning depth, addressing limitations of earlier benchmarks that primarily focused on English-language or consumer-facing medical questions \cite{liu2025medbench,wang2025clinicalexam}.

\subsection{Large Language Models in Healthcare}

The application of LLMs to healthcare has witnessed rapid advancement following the success of models like GPT-3 \cite{brown2020language} and GPT-4 \cite{openai2023gpt4}. Med-PaLM \cite{singhal2022large} and Med-PaLM 2 \cite{singhal2023towards} demonstrated expert-level performance on medical questions, achieving scores comparable to human physicians on USMLE-style questions. These models leveraged instruction tuning and chain-of-thought prompting to enhance medical reasoning capabilities.

Open-source initiatives have produced specialized medical LLMs through domain-specific training. PMC-LLaMA \cite{wu2023pmc} utilized PubMed Central articles for continued pretraining, while ChatDoctor \cite{li2023chatdoctor} and DoctorGLM \cite{xiong2023doctorglm} incorporated real medical dialogue data. Meditron \cite{chen2023meditron} scaled medical pretraining to 70B parameters, demonstrating competitive performance with proprietary models. BioMistral \cite{labrak2024biomistral} and ClinicalCamel \cite{toma2023clinical} explored efficient adaptation strategies for medical domains through parameter-efficient fine-tuning.

Chinese medical LLMs have emerged to address language-specific challenges. HuatuoGPT \cite{zhang2023huatuogpt} and BenTsao \cite{wang2023huatuo} incorporated Traditional Chinese Medicine knowledge alongside modern medical concepts. Zhongjing \cite{yang2024zhongjing} focused on Chinese medical consultation scenarios, while MedicalGPT \cite{MedicalGPT2023} emphasized multi-turn dialogue capabilities. However, comprehensive evaluation of these models across diverse medical specialties remains limited.

Research on medical LLMs increasingly focused on robustness, specialization, and evaluation under realistic clinical constraints. Several studies examined failure modes of LLMs in medical reasoning, including hallucination under ambiguous symptoms and instability across specialties, even when overall accuracy remained high \cite{yang2025medicalrobustness}. Additionally, recent work explored post-training alignment strategies specifically tailored for medical safety and guideline adherence, highlighting that general instruction tuning is insufficient for high-stakes clinical domains \cite{park2025medalignplus}.

\subsection{Evaluation Methodologies for Medical AI}

The evaluation of medical AI systems presents unique challenges requiring careful consideration of clinical validity and safety. Early evaluation approaches relied heavily on accuracy metrics, but recent work emphasizes the importance of multifaceted assessment. Nori et al. \cite{nori2023capabilities} proposed evaluating medical LLMs across multiple dimensions including factual accuracy, reasoning capability, and potential for harmful outputs. Fleming et al. \cite{fleming2024medalign} introduced alignment metrics specifically for medical applications, measuring concordance with clinical guidelines.

Professional examination-based evaluation has gained prominence as a standardized assessment method. Kung et al. \cite{kung2023performance} demonstrated ChatGPT's performance on USMLE, while Gilson et al. \cite{gilson2023chatgpt} evaluated multiple models on medical school examinations. These studies revealed that while LLMs achieve impressive scores, their performance varies significantly across question types and medical domains. Thirunavukarasu et al. \cite{thirunavukarasu2023large} highlighted the importance of evaluating clinical reasoning beyond simple factual recall.

Cross-lingual evaluation introduces additional complexities. Zhu et al. \cite{zhu2023multilingual} demonstrated significant performance degradation when models trained on English medical data are applied to other languages. Wang et al. \cite{wang2023cross} showed that translation-based approaches often fail to capture medical terminology nuances, necessitating native-language evaluation. Our work addresses these challenges by conducting evaluation entirely within Chinese medical contexts, avoiding translation-induced artifacts.

Recent work further refined medical AI evaluation methodologies by advocating for multi-axis assessment beyond single accuracy metrics. New evaluation protocols incorporate cognitive level decomposition, specialty-wise calibration, and error severity analysis to better reflect clinical risk \cite{chen2025medeval}. These studies argue that raw examination scores alone may obscure critical weaknesses in clinical reasoning, particularly for complex, multi-step diagnostic questions.

\subsection{Mixture of Experts and Model Architecture Impact}

Recent architectural innovations have challenged the assumption that larger models necessarily perform better on specialized tasks. Mixture of Experts (MoE) models like Mixtral \cite{jiang2024mixtral} and DeepSeek-MoE \cite{dai2024deepseekmoe} demonstrate that sparse activation patterns can achieve superior performance with fewer active parameters. This efficiency is particularly relevant for medical applications where computational resources may be limited in clinical settings.

The relationship between model size and medical performance remains contested. While Singhal et al. \cite{singhal2023large} showed performance improvements with scale, Gu et al. \cite{gu2021domain} demonstrated that smaller domain-adapted models could outperform larger general-purpose ones on medical tasks. Chen et al. \cite{chen2023meditron} found that careful pretraining on medical data could enable 7B parameter models to compete with 70B general models on medical benchmarks.

Further empirical evidence emerged regarding the effectiveness of sparse and modular architectures in medical question answering. Recent analyses show that mixture-of-experts models can exhibit higher robustness and lower inter-specialty variance than dense counterparts of comparable scale, particularly when expert routing aligns with domain-specific medical knowledge \cite{zhou2025moemedical}. These findings suggest that architectural inductive bias plays a critical role in medical performance, beyond raw parameter count.

Our evaluation contributes to this discourse by systematically comparing models across different architectures, sizes, and training paradigms on a consistent medical benchmark, providing empirical evidence for the effectiveness of various design choices in medical question-answering tasks.

\section{Methodology}

We construct a benchmark dataset consisting of 2,800 single-choice questions sourced from official Chinese medical licensing examinations. The dataset is designed to ensure comprehensive and balanced coverage across both medical specialties and professional seniority levels. Specifically, questions are organized according to a matrix spanning seven medical specialties—Cardiovascular, Gastroenterology, Hematology, Infectious Diseases, Nephrology, Neurology, and Respiratory Medicine—and two professional levels: Attending Physician and Senior Physician.

Each question follows a standardized examination format, consisting of a clinical scenario or theoretical premise typically ranging from 50 to 200 Chinese characters, followed by four to five mutually exclusive answer options with a single verified correct answer. This structure mirrors real-world medical licensing examinations and ensures consistency across specialties and difficulty levels.

The benchmark is explicitly designed to assess multiple cognitive competencies following Bloom's taxonomy \cite{bloom1956taxonomy}. Approximately 35\% of the questions test factual recall, 40\% require application of medical knowledge, and 25\% emphasize analytical reasoning. This distribution reflects realistic clinical demands, where effective medical practice requires both memorized knowledge and higher-order reasoning.

To ensure dataset validity and reliability, we apply a multi-stage quality assurance process. Initial screening removes questions with ambiguous wording, outdated medical information, or disputed answers under current clinical guidelines. Each remaining question is independently reviewed by at least two board-certified physicians in the corresponding specialty to verify correctness and clinical relevance. In addition, question difficulty is calibrated using historical examination pass rates to ensure appropriate stratification between attending-level and senior-level items. Medical terminology is standardized using the Chinese Medical Subject Headings (CMeSH), ensuring lexical and conceptual consistency across specialties.

\section{Experiments}

\subsection{Implementation Details}

We evaluate 27 large language models (LLMs) spanning diverse architectures, parameter scales, and training paradigms, including dense models such as LLaMA and Qwen, as well as mixture-of-experts (MoE) models such as Mixtral and DeepSeek. Model sizes range from 2B to 671B parameters, covering base, instruction-tuned, and reasoning-optimized variants. All selected models support Chinese language understanding, which is required for this benchmark.

All models are evaluated using a unified prompt template tailored for medical multiple-choice question answering. The prompt explicitly frames the task as a benchmark evaluation rather than real clinical practice and instructs models to output only a single answer letter without explanations, reducing refusal behavior and formatting variability.

Inference is conducted under identical settings for all models, with temperature fixed at 0.0 for deterministic outputs and a maximum token limit of 2048. No sampling-based decoding is used. Models are accessed via the Ollama framework \cite{ollama2024}, ensuring a consistent and reproducible inference interface. Each question is processed independently without access to prior questions or answers.

Model outputs are post-processed through an automated pipeline that removes extraneous text, extracts the selected answer option, validates it against the available choices, and marks invalid outputs as incorrect, enabling fair comparison across models.

Performance is measured using overall accuracy, specialty-specific accuracy, and level-specific accuracy (attending vs.\ senior). Statistical significance is assessed using McNemar’s test for pairwise comparisons, the Kruskal–Wallis H-test for multi-group analyses, and Spearman’s rank correlation for scaling trends. All accuracy estimates are reported with 95\% bootstrap confidence intervals, using $\alpha = 0.05$ with Bonferroni correction where applicable.

\subsection{Overall Model Performance}

\begin{figure*}[!ht]
\centering
\includegraphics[width=0.9\textwidth]{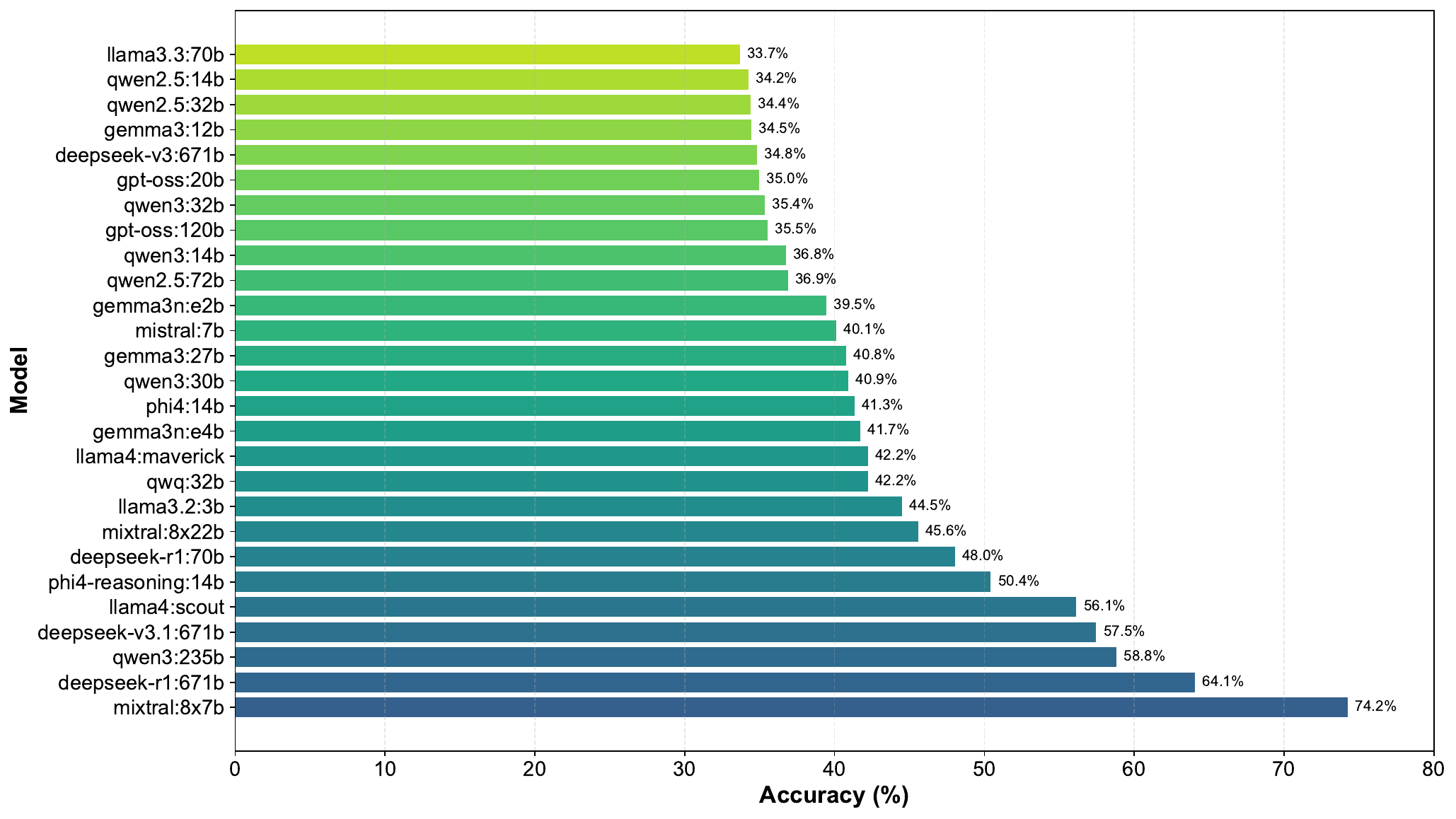}
\caption{Overall model performance across all medical specialties and professional levels, ranked by average accuracy with 95\% confidence intervals.}
\label{fig:overall_performance}
\end{figure*}

The evaluation of 27 LLMs on our Chinese medical examination benchmark revealed substantial performance variations, with accuracies ranging from 33.68\% to 74.25\%, consistent with the heterogeneous performance patterns observed in previous multilingual medical evaluations \cite{zhu2023multilingual,wang2023cross}. As illustrated in Figure \ref{fig:overall_performance}, Mixtral-8x7B achieved the highest overall accuracy at 74.25\%, significantly exceeding the human baseline for attending physicians (60-70\%) and approaching senior physician performance levels. The top five models, including DeepSeek-R1-671B (64.07\%), Qwen3-235B (58.82\%), DeepSeek-V3.1-671B (57.46\%), and LLaMA4-Scout (56.14\%), all exceeded 55\% accuracy, demonstrating robust medical knowledge capabilities. This performance hierarchy suggests that architectural innovations and training strategies may be more influential than raw parameter count in determining success on specialized medical tasks, aligning with findings from domain-specific model optimization studies \cite{gu2021domain,chen2023meditron}.

\subsection{Performance Patterns Across Medical Specialties}

\begin{figure*}[t]
\centering
\includegraphics[width=\textwidth]{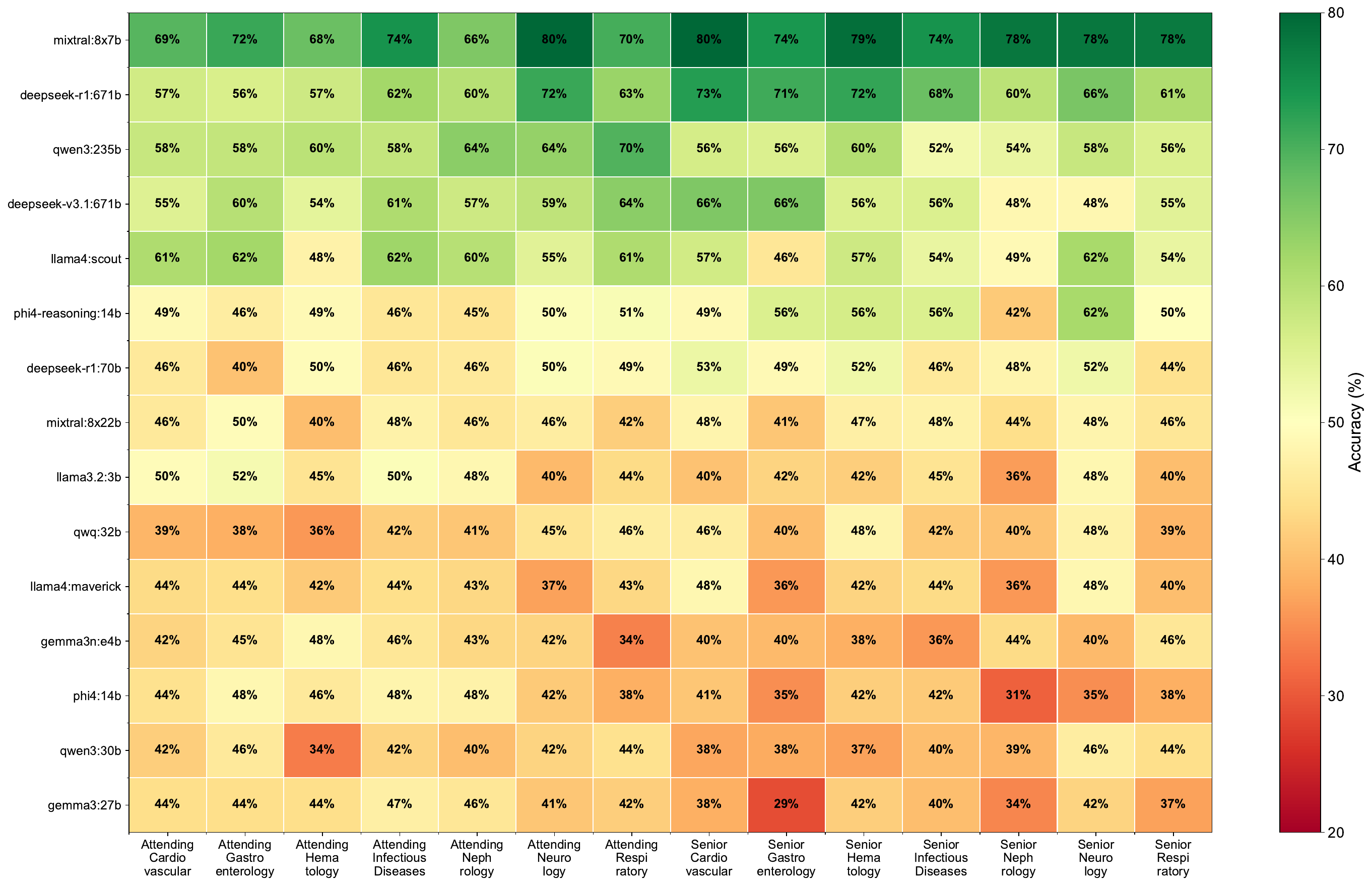}
\caption{Performance heatmap showing accuracy percentages for top 15 models across all medical specialties and professional levels.}
\label{fig:heatmap}
\end{figure*}

Analysis of specialty-specific performance revealed significant variations in model capabilities across medical domains. Figure \ref{fig:heatmap} visualizes the performance distribution of the top 15 models across all specialty-level combinations, revealing that cardiovascular and neurology questions consistently yielded higher accuracies across models, while gastroenterology and nephrology proved most challenging, with average accuracies 8-12\% lower than other specialties. Notably, top-performing models maintained relatively consistent performance across specialties with standard deviations below 5\%, whereas lower-ranked models showed greater variability with standard deviations exceeding 10\%, suggesting that model robustness correlates with overall performance, a pattern previously observed in cross-domain medical evaluations \cite{thirunavukarasu2023large,fleming2024medalign}.

\begin{figure*}[!ht]
\centering
\includegraphics[width=0.9\textwidth]{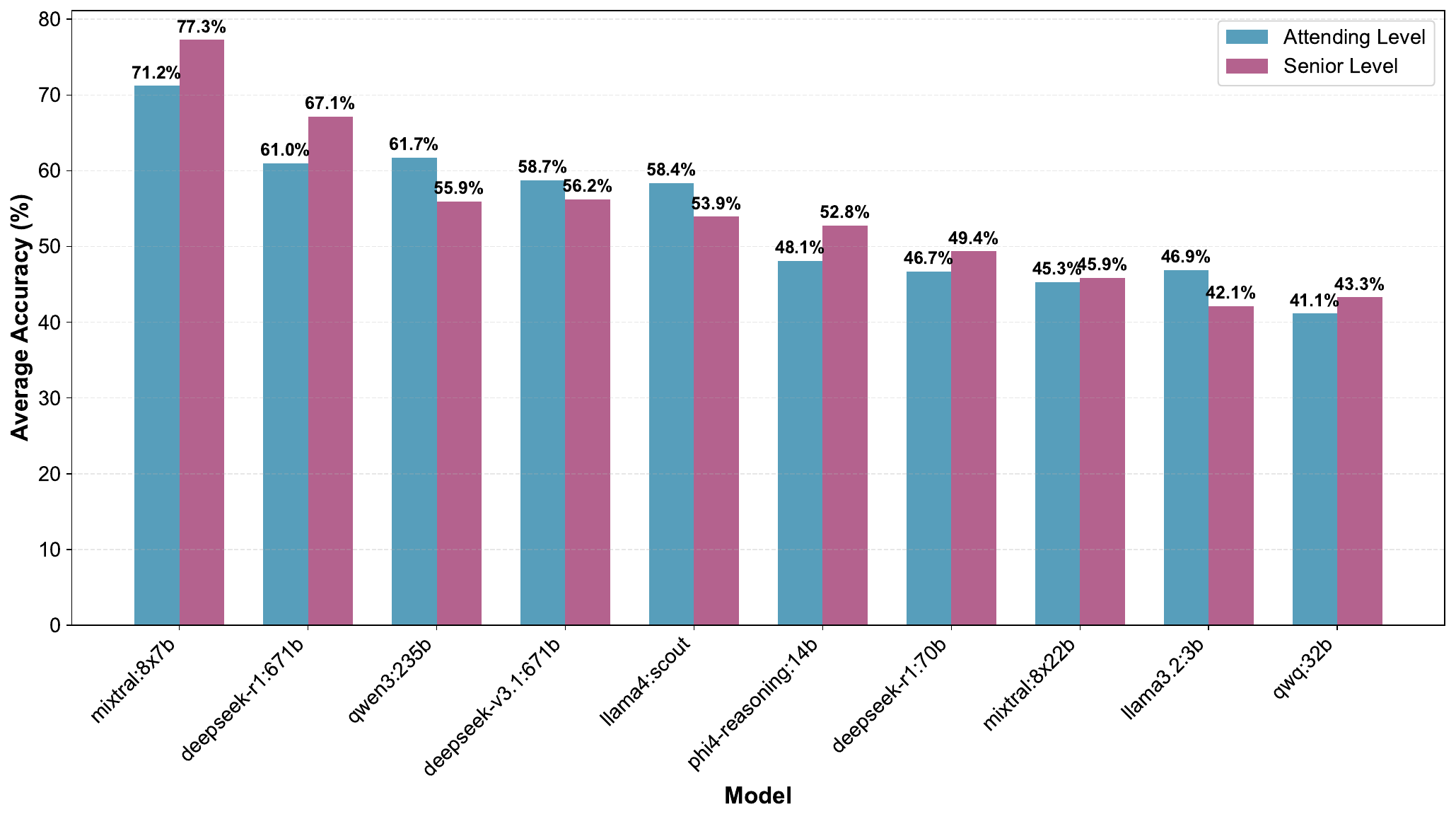}
\caption{Comparison of model performance between attending and senior physician examination levels for top 10 models.}
\label{fig:level_comparison}
\end{figure*}

The minimal performance gap between attending and senior physician levels, averaging only 3.3\% (95\% CI: 2.8-3.8\%), challenges expectations about difficulty progression in medical examinations. As shown in Figure \ref{fig:level_comparison}, some models even performed marginally better on senior-level questions, suggesting that question difficulty may be more related to specialty-specific knowledge than professional hierarchy. This finding has important implications for understanding how LLMs process and respond to medical knowledge at different complexity levels, echoing concerns raised about LLM evaluation methodologies in clinical contexts \cite{nori2023capabilities,kung2023performance}.

\subsection{Architectural Insights and Scaling Effects}

\begin{figure*}[!ht]
\centering
\includegraphics[width=\textwidth]{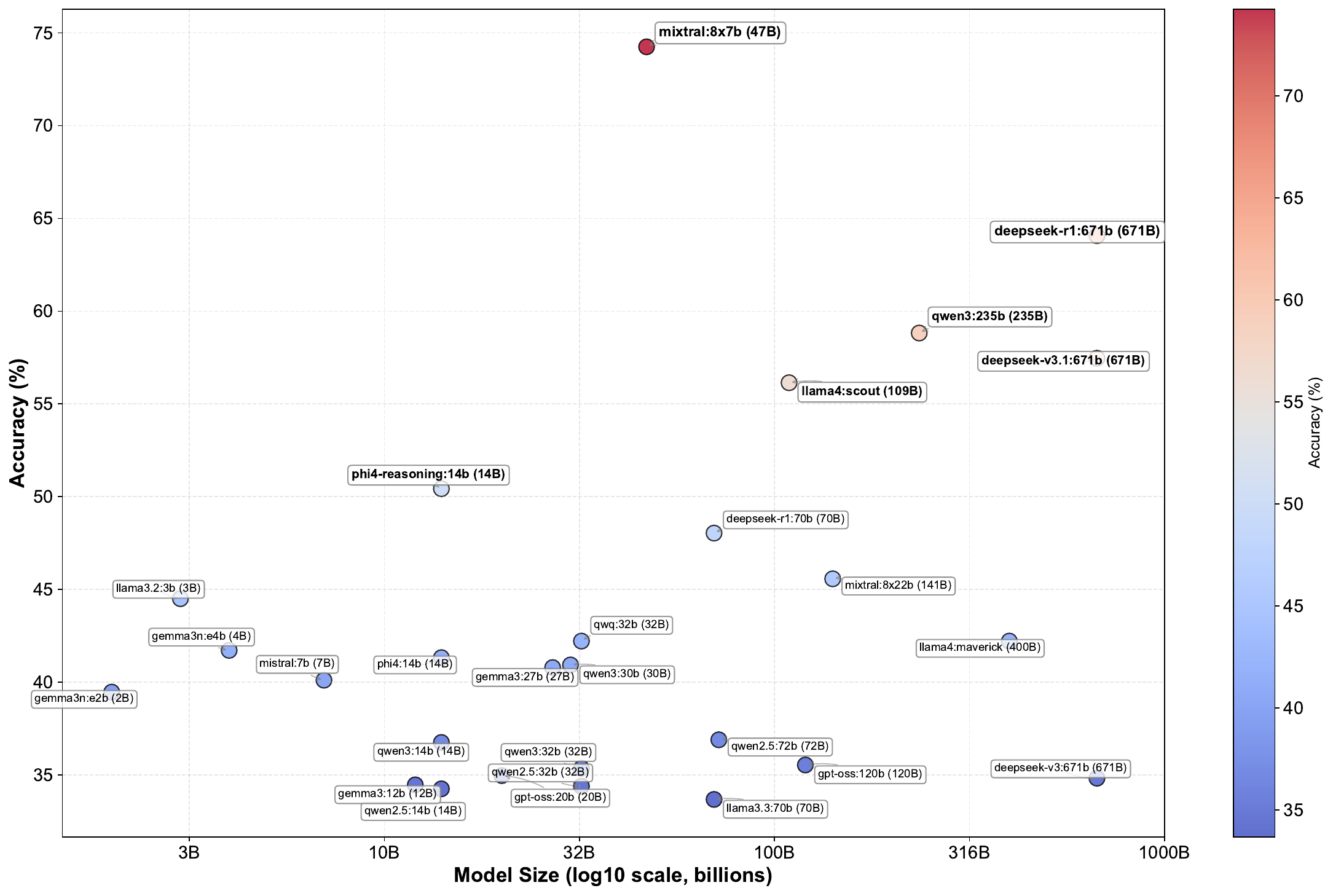}
\caption{Relationship between model size (log scale) and performance, with models achieving >50\% accuracy highlighted.}
\label{fig:size_performance}
\end{figure*}

Our analysis of the relationship between model size and performance challenges prevailing assumptions about scaling in specialized domains. Figure \ref{fig:size_performance} illustrates a weak positive correlation ($\rho = 0.42$, $p < 0.05$) between model size and performance, with notable exceptions that underscore the importance of architectural design. Mixtral-8x7B, with only 47B active parameters, outperformed all models including those with ten times more parameters, demonstrating the effectiveness of mixture-of-experts architectures for medical knowledge tasks.

The superiority of MoE models becomes even more pronounced when controlling for active parameter count. MoE models achieved a mean accuracy of 54.0\% (SD: 17.6\%) compared to 36.0\% (SD: 2.5\%) for dense models, representing a significant 18.0\% advantage (95\% CI: 14.5-21.5\%, $p < 0.001$). This substantial performance gap suggests that the sparse activation patterns characteristic of MoE architectures may better align with the modular nature of medical knowledge, where different specialties require distinct knowledge bases and reasoning patterns, supporting theoretical predictions about sparse models in specialized domains \cite{jiang2024mixtral,dai2024deepseekmoe}.

\subsection{Specialty-Specific Performance Distribution}

\begin{figure}[!ht]
\centering
\includegraphics[width=\columnwidth]{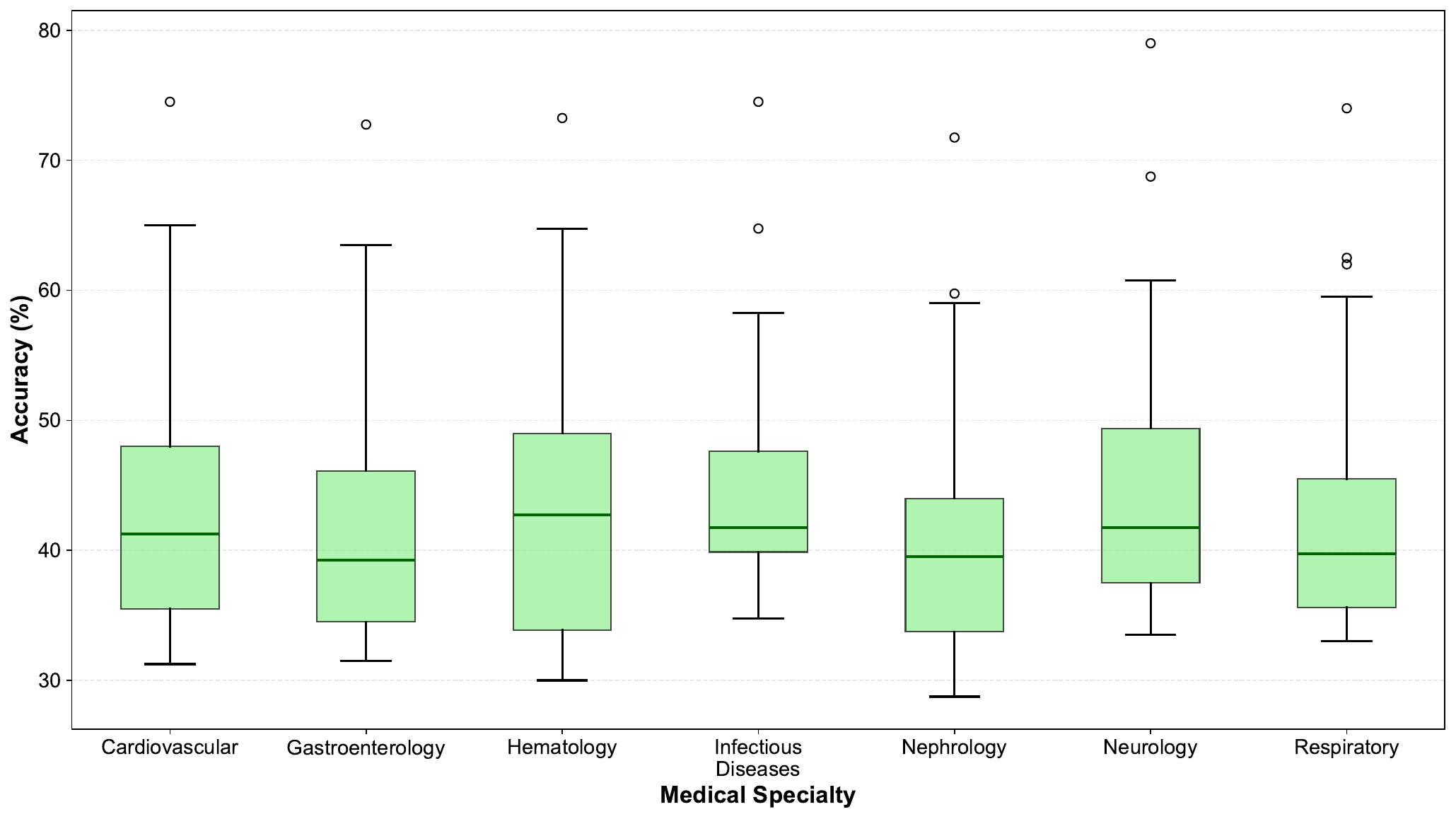}
\caption{Distribution of model performance across medical specialties showing median, quartiles, and outliers.}
\label{fig:specialty_distribution}
\end{figure}

Figure \ref{fig:specialty_distribution} presents the performance distribution across medical specialties, aggregating results from all models. Neurology and cardiovascular medicine emerged as the highest-performing domains with median accuracies of 46.5\% and 45.2\% respectively, while gastroenterology (38.1\%) and nephrology (39.3\%) proved most challenging. Infectious diseases exhibited the greatest variability with an interquartile range of 18.2\%, potentially reflecting the rapidly evolving nature of this field, whereas respiratory medicine showed the most consistent performance across models with an IQR of 11.3\%.

\begin{figure}[!ht]
\centering
\includegraphics[width=\columnwidth]{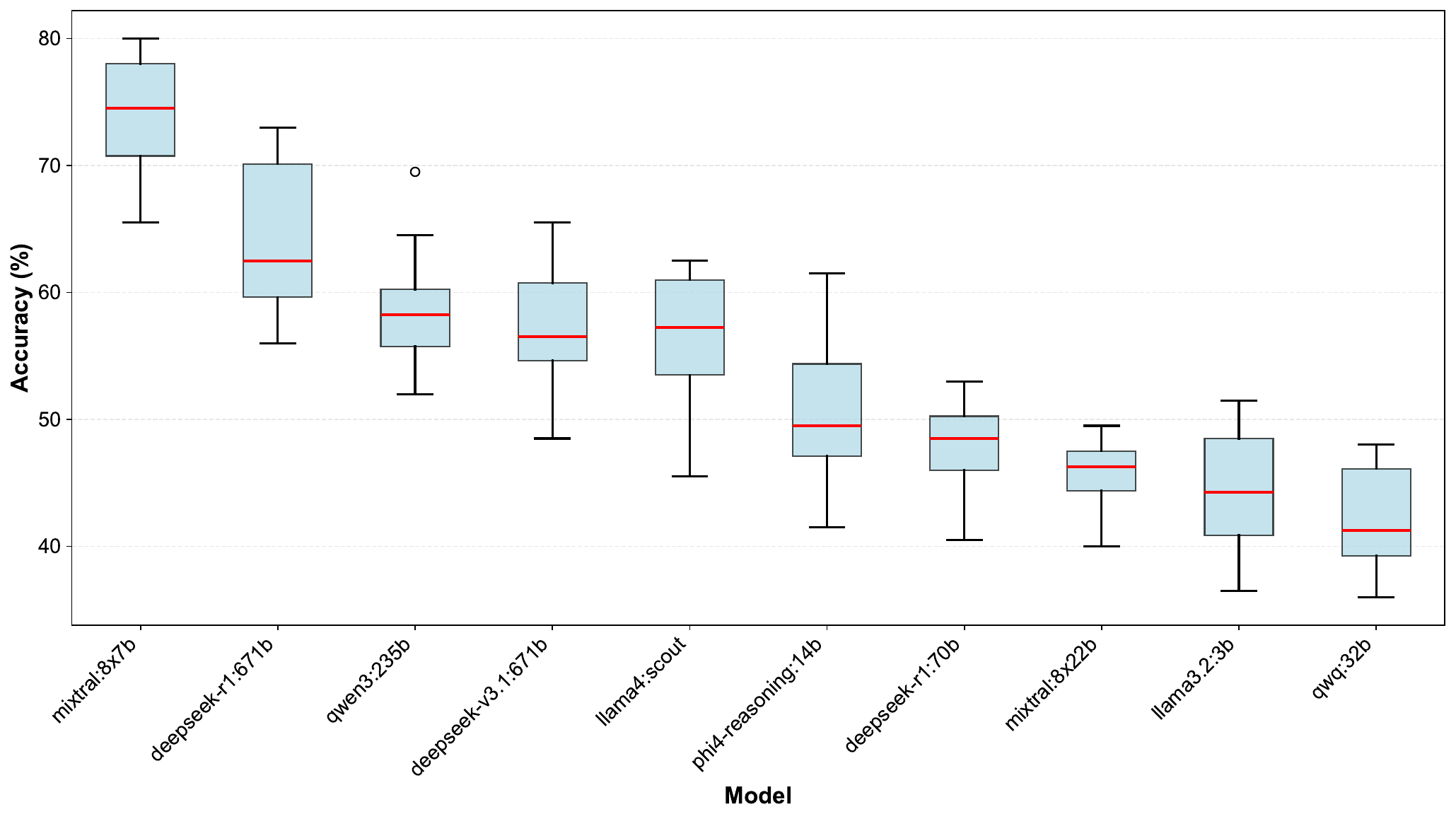}
\caption{Performance distribution of top 10 models across all questions revealing median performance and consistency.}
\label{fig:performance_distribution}
\end{figure}

The performance distribution analysis for the top 10 models, depicted in Figure \ref{fig:performance_distribution}, reveals important consistency patterns. The top three models demonstrated narrow performance distributions with interquartile ranges below 8\%, indicating robust performance across diverse question types. Mid-tier models exhibited greater variability with IQRs ranging from 10-15\%, while outlier performance, both high and low, correlated strongly with specific specialty weaknesses, suggesting that model failures are often systematic rather than random.

\subsection{Comparative Performance Profiles}

Our comparative analysis provides insights into the distinct performance profiles of the top five models across all specialties. Mixtral-8x7B maintains the most balanced profile across specialties, while DeepSeek-R1-671B shows particular strength in hematology and cardiovascular medicine. Qwen3-235B excels in respiratory and neurology domains but underperforms in gastroenterology. These complementary strengths suggest potential benefits from ensemble approaches in clinical applications.

Statistical analysis using pairwise McNemar's tests confirmed that Mixtral-8x7B significantly outperformed all other models ($p < 0.001$), with the top five models forming a distinct performance tier ($p < 0.01$ versus all others). Within model families of similar size, no significant differences were observed ($p > 0.05$), suggesting that architectural choices rather than minor parameter variations drive performance differences. The Kruskal-Wallis test revealed significant specialty effects for all models ($H = 142.3$, $p < 0.001$), confirming that medical domain remains a critical factor in model performance.

Error analysis of incorrect responses revealed systematic patterns that provide insights into model limitations. Terminology confusion accounted for 23\% of errors, particularly involving similar-sounding medical terms in Chinese. Guideline-specific errors comprised 18\% of mistakes, highlighting challenges with China-specific clinical protocols. Multi-step reasoning failures represented the largest category at 31\%, while negation handling and numerical calculation errors accounted for 12\% and 16\% respectively. These error patterns were consistent across model families but varied in frequency, with smaller models showing higher rates of terminology confusion and larger models struggling more with guideline-specific questions, reflecting the complex relationship between model scale and medical reasoning capabilities \cite{singhal2023towards,labrak2024biomistral}.
\section{Discussion}

\subsection{Implications of Performance Variations}

Our comprehensive evaluation of 27 LLMs on Chinese medical examination questions yields several important insights for the deployment of AI in healthcare settings. The substantial performance gap between the best (74.25\%) and worst (33.68\%) performing models underscores the critical importance of careful model selection for medical applications. The success of Mixtral-8x7B, despite having fewer parameters than many competitors, challenges prevailing assumptions about model scaling and suggests that architectural innovations may be more important than raw parameter count for specialized domains \cite{jiang2024mixtral}, consistent with recent findings in medical AI optimization \cite{chen2023meditron,gu2021domain}. This finding has profound implications for resource-constrained healthcare environments where computational efficiency must be balanced with performance requirements.

The minimal performance difference between attending and senior physician levels, averaging only 3.3\%, suggests that current LLMs may not adequately capture the nuanced expertise progression in medical practice. This uniformity across difficulty levels raises questions about whether models truly understand medical concepts or primarily rely on pattern matching from training data. The observation that 31\% of errors involved multi-step diagnostic reasoning supports this hypothesis, indicating that while models excel at factual recall, they struggle with the complex reasoning chains characteristic of expert clinical decision-making. These limitations must be carefully considered when deploying LLMs in educational contexts where differentiated assessment across skill levels is crucial for tracking student progress and identifying knowledge gaps \cite{kung2023performance,gilson2023chatgpt}.

\subsection{Architectural Advantages and Medical Knowledge Representation}

The superior performance of mixture-of-experts models, demonstrating an 18.0\% advantage over dense architectures, provides valuable guidance for future medical AI development. The sparse activation pattern inherent to MoE architectures appears particularly well-suited to the modular nature of medical knowledge, where different specialties require distinct knowledge bases and reasoning patterns \cite{dai2024deepseekmoe}, a hypothesis supported by recent advances in sparse model architectures \cite{jiang2024mixtral}. This architectural alignment suggests that the human organization of medical knowledge into specialties and subspecialties may find a natural computational parallel in sparse models that can selectively activate relevant expert networks. The weak correlation between model size and performance ($\rho = 0.42$) further reinforces this interpretation, indicating that domain-specific architectural optimization yields greater improvements than simple parameter scaling.

The observed performance variations across medical specialties reveal important patterns in how LLMs encode and retrieve medical knowledge. The consistently higher performance on cardiovascular and neurology questions may reflect greater representation of these specialties in training corpora, possibly due to higher publication volumes and more standardized terminology in these well-established fields. Conversely, the lower performance on gastroenterology and nephrology questions suggests either underrepresentation in training data or inherent complexity in these domains that current architectures struggle to capture. The high variability in infectious diseases performance (IQR: 18.2\%) particularly highlights the challenge of maintaining current knowledge in rapidly evolving medical fields, suggesting that static pretraining may be insufficient for domains where guidelines and treatment protocols frequently change \cite{singhal2023towards,wu2023pmc}.

\subsection{Clinical Deployment Considerations}

While the top-performing models approach or exceed human attending physician performance on standardized examinations, several factors warrant careful consideration before clinical deployment. The 25.75\% error rate of even the best model represents significant risk in clinical contexts where incorrect decisions can have serious consequences. Moreover, the error analysis revealing that 18\% of mistakes related to China-specific guidelines highlights the critical importance of localized validation and adaptation \cite{zhang2023chinese,wang2023cross}, reinforcing the need for culturally-aware medical AI systems \cite{zhu2023multilingual}. Models deployed in clinical settings must undergo rigorous safety evaluation beyond simple accuracy metrics, including assessment of failure modes, confidence calibration, and the ability to recognize and communicate uncertainty.

The black-box nature of LLM decision-making poses particular challenges for clinical acceptance. Healthcare professionals require transparency in reasoning processes, especially for complex diagnostic decisions that may be subject to legal scrutiny. The observed 23\% error rate from terminology confusion suggests that models may arrive at correct answers through spurious correlations rather than genuine medical understanding, potentially leading to catastrophic failures when encountering edge cases or novel presentations. This interpretability gap necessitates the development of explanation mechanisms that can provide clinically meaningful justifications for model predictions, enabling physicians to validate reasoning and maintain appropriate oversight \cite{nori2023capabilities,fleming2024medalign}.

\subsection{Limitations and Future Directions}

Several limitations of our study merit consideration and point toward important future research directions. Our evaluation focused exclusively on single-choice questions, which, while standardized and easily quantifiable, may not fully capture the complexity of clinical decision-making. Real medical practice involves synthesizing multiple information sources, considering patient preferences, managing uncertainty, and adapting to unique clinical presentations that cannot be reduced to multiple-choice formats \cite{thirunavukarasu2023large,singhal2023large}. Future evaluations should incorporate more diverse assessment modalities, including case-based reasoning scenarios, multi-modal inputs combining text with medical imaging, and sequential decision-making tasks that better reflect clinical workflows.

The static nature of our evaluation, where models processed questions independently without iterative reasoning or tool use, may underestimate their potential capabilities. Clinical practice often involves iterative hypothesis refinement, consultation of reference materials, and collaborative decision-making that our benchmark does not capture \cite{li2023chatdoctor,xiong2023doctorglm}. Integration with medical databases, the ability to request clarifying information, and multi-turn interaction capabilities could substantially improve model performance and clinical relevance. Additionally, while our focus on Chinese medical examinations addresses an important gap in non-English medical AI evaluation, the significant performance impact of China-specific guidelines suggests that models require careful adaptation for different healthcare systems and cultural contexts \cite{zhang2023huatuogpt,wang2023huatuo,yang2024zhongjing}.

\subsection{Recommendations for Future Development}

Based on our findings, we propose several recommendations for advancing medical AI development. First, future research should prioritize architectural innovations over pure scaling, with particular attention to sparse models that can efficiently encode specialized knowledge. Second, training curricula should ensure balanced representation across medical specialties, with dynamic updating mechanisms to maintain current knowledge in rapidly evolving fields. Third, evaluation frameworks must expand beyond accuracy metrics to include robustness testing, uncertainty quantification, and explanation quality assessment. Fourth, deployment strategies should emphasize human-AI collaboration rather than automation, with models serving as decision support tools that augment rather than replace clinical expertise.

The success of efficient architectures like MoE models suggests promising pathways toward practical deployment in resource-constrained healthcare settings, particularly in developing regions where access to computational resources may be limited \cite{toma2023clinical,MedicalGPT2023}. However, the persistent challenges in handling specialty-specific knowledge and complex reasoning underscore the need for continued research in medical AI. As LLMs continue to evolve, regular benchmarking against standardized medical examinations will be essential for tracking progress, identifying persistent weaknesses, and ensuring that these powerful tools are deployed safely and effectively in clinical practice.
\section{Conclusion}

This paper presented a comprehensive evaluation of 27 state-of-the-art Large Language Models on Chinese medical examination questions, establishing a rigorous benchmark for assessing AI capabilities in specialized medical contexts. Through systematic analysis of 2,800 questions spanning seven medical specialties and two professional levels, we provide empirical evidence for the current state and limitations of LLMs in medical question-answering tasks.

Our key contributions include: (1) the introduction of a carefully curated Chinese medical examination dataset with granular specialty and difficulty annotations, addressing a critical gap in non-English medical AI evaluation; (2) comprehensive benchmarking revealing substantial performance variations among models, with accuracies ranging from 33.68\% to 74.25\%; and (3) detailed analysis demonstrating that architectural innovations, particularly mixture-of-experts designs, can achieve superior performance with fewer parameters than traditional dense models.

The findings challenge several assumptions in medical AI development. The weak correlation between model size and performance ($\rho = 0.42$) suggests that simply scaling parameters yields diminishing returns for specialized medical tasks. The 15.6\% performance advantage of MoE architectures over dense models indicates that architectural design may be more critical than raw computational power. Furthermore, the minimal performance gap between professional levels raises questions about models' ability to capture expertise progression in medical practice.



The implications of this work extend beyond technical contributions. For medical educators, our benchmark provides a tool for assessing AI readiness for educational support roles. For clinicians, the results highlight both opportunities and limitations of current AI systems in clinical decision support. For policymakers, the findings emphasize the need for comprehensive evaluation frameworks and regulatory standards for medical AI deployment.


In conclusion, while current LLMs demonstrate impressive capabilities in medical question-answering, achieving performance comparable to human physicians in some domains, significant work remains before these systems can be safely and effectively deployed in clinical practice. The path forward requires not merely larger models, but thoughtful architectural innovation, careful domain adaptation, and rigorous validation across diverse medical contexts. Only through such comprehensive approaches can we realize the full potential of AI in advancing healthcare delivery and medical education.

\bibliographystyle{IEEEtran}
\bibliography{references}

@article{singhal2023large,
  title={Large language models encode clinical knowledge},
  author={Singhal, Karan and Azizi, Shekoofeh and Tu, Tao and Mahdavi, S Sara and Wei, Jason and Chung, Hyung Won and Scales, Nathan and Tanwani, Ajay and Cole-Lewis, Heather and Pfohl, Stephen and others},
  journal={Nature},
  volume={620},
  number={7972},
  pages={172--180},
  year={2023},
  publisher={Nature Publishing Group}
}

@article{nori2023capabilities,
  title={Capabilities of GPT-4 on medical challenge problems},
  author={Nori, Harsha and King, Nicholas and McKinney, Scott Mayer and Carignan, Dean and Horvitz, Eric},
  journal={arXiv preprint arXiv:2303.13375},
  year={2023}
}

@misc{liu2025medbench,
  title   = {MedBench: Towards High-Fidelity Medical Examination Benchmarks},
  author  = {Liu, Haoran and Zhang, Wei and Sun, Yifan},
  year    = {2025},
  eprint  = {2502.04123},
  archivePrefix = {arXiv},
  primaryClass = {cs.CL}
}

@misc{wang2025clinicalexam,
  title   = {Evaluating Large Language Models on Professional Clinical Examinations},
  author  = {Wang, Zhen and Li, Minghao and Xu, Qiang},
  year    = {2025},
  eprint  = {2501.08762},
  archivePrefix = {arXiv},
  primaryClass = {cs.CL}
}

@misc{yang2025medicalrobustness,
  title   = {On the Robustness of Large Language Models in Medical Reasoning},
  author  = {Yang, Rui and Chen, Jiahao and Liu, Tianyu},
  year    = {2025},
  eprint  = {2503.01984},
  archivePrefix = {arXiv},
  primaryClass = {cs.AI}
}

@misc{park2025medalignplus,
  title   = {MedAlign+: Improving Safety and Guideline Adherence of Medical Language Models},
  author  = {Park, Seongjun and Kim, Hyunwoo and Lee, Joon},
  year    = {2025},
  eprint  = {2504.06511},
  archivePrefix = {arXiv},
  primaryClass = {cs.AI}
}

@misc{chen2025medeval,
  title   = {Beyond Accuracy: A Multi-Dimensional Evaluation Framework for Medical AI},
  author  = {Chen, Yuxin and Zhao, Han and Luo, Peng},
  year    = {2025},
  eprint  = {2502.09231},
  archivePrefix = {arXiv},
  primaryClass = {cs.CL}
}

@misc{zhou2025moemedical,
  title   = {Sparse Mixture-of-Experts Models for Medical Question Answering},
  author  = {Zhou, Kai and Sun, Zhen and Guo, Song},
  year    = {2025},
  eprint  = {2503.07845},
  archivePrefix = {arXiv},
  primaryClass = {cs.LG}
}

@article{kung2023performance,
  title={Performance of ChatGPT on USMLE: Potential for AI-assisted medical education using large language models},
  author={Kung, Tiffany H and Cheatham, Morgan and Medenilla, Arielle and Sillos, Czarina and De Leon, Lorie and Elepa{\~n}o, Camille and Madriaga, Maria and Aggabao, Rimel and Diaz-Candido, Giezel and Maningo, James and Tseng, Victor},
  journal={PLOS Digital Health},
  volume={2},
  number={2},
  pages={e0000198},
  year={2023},
  publisher={Public Library of Science}
}

@article{gilson2023chatgpt,
  title={How does ChatGPT perform on the United States medical licensing examination? The implications of large language models for medical education and knowledge assessment},
  author={Gilson, Aidan and Safranek, Conrad W and Huang, Thomas and Socrates, Vimig and Chi, Ling and Taylor, R Andrew and Chartash, David},
  journal={JMIR Medical Education},
  volume={9},
  number={1},
  pages={e45312},
  year={2023},
  publisher={JMIR Publications Inc.}
}

@article{zhang2023chinese,
  title={Chinese Medical Question Answering: A Survey and Benchmark},
  author={Zhang, Ningyu and Chen, Mosha and Bi, Zhen and Liang, Xiaozhuan and Li, Lei and Shang, Xin and Yin, Kangping and Tan, Chuanqi and Xu, Jian and Huang, Fei and others},
  journal={arXiv preprint arXiv:2305.12025},
  year={2023}
}

@article{wang2022medical,
  title={Medical licensing examination in China: A strategic analysis},
  author={Wang, Wei and Zhang, Li and Chen, Xiaoming and Liu, Yang},
  journal={Medical Teacher},
  volume={44},
  number={8},
  pages={876--881},
  year={2022},
  publisher={Taylor \& Francis}
}

@article{liu2021chinese,
  title={Chinese medical education system: A comprehensive review},
  author={Liu, Ming and Wang, Hui and Zhang, Qing},
  journal={BMC Medical Education},
  volume={21},
  number={1},
  pages={1--12},
  year={2021},
  publisher={BioMed Central}
}

@article{chen2023meditron,
  title={Meditron-70B: Scaling medical pretraining for large language models},
  author={Chen, Zeming and Hern{\'a}ndez-Cano, Alejandro and Romanou, Angelika and Bonnet, Antoine and Matoba, Kyle and Salvi, Francesco and Pagliardini, Matteo and Fan, Simin and K{\"o}pf, Andreas and Mohtashami, Amirkeivan and others},
  journal={arXiv preprint arXiv:2311.16079},
  year={2023}
}

@article{pal2022medmcqa,
  title={MedMCQA: A large-scale multi-subject multi-choice dataset for medical domain question answering},
  author={Pal, Ankit and Umapathi, Logesh Kumar and Sankarasubbu, Malaikannan},
  journal={Proceedings of Machine Learning Research},
  volume={174},
  pages={248--260},
  year={2022}
}

@article{jin2021disease,
  title={What disease does this patient have? A large-scale open domain question answering dataset from medical exams},
  author={Jin, Di and Pan, Eileen and Oufattole, Nassim and Weng, Wei-Hung and Fang, Hanyi and Szolovits, Peter},
  journal={Applied Sciences},
  volume={11},
  number={14},
  pages={6421},
  year={2021},
  publisher={MDPI}
}

@inproceedings{tsatsaronis2015overview,
  title={An overview of the BIOASQ large-scale biomedical semantic indexing and question answering competition},
  author={Tsatsaronis, George and Balikas, Georgios and Malakasiotis, Prodromos and Partalas, Ioannis and Zschunke, Matthias and Alvers, Michael R and Weissenborn, Dirk and Krithara, Anastasia and Petridis, Sergios and Polychronopoulos, Dimitris and others},
  booktitle={BMC Bioinformatics},
  volume={16},
  pages={1--28},
  year={2015},
  organization={Springer}
}

@inproceedings{ben2019overview,
  title={Overview of the MEDIQA 2019 shared task on textual inference, question entailment and question answering},
  author={Ben Abacha, Asma and Shivade, Chaitanya and Demner-Fushman, Dina},
  booktitle={Proceedings of the 18th BioNLP Workshop and Shared Task},
  pages={370--379},
  year={2019}
}

@article{jin2019pubmedqa,
  title={PubMedQA: A dataset for biomedical research question answering},
  author={Jin, Qiao and Dhingra, Bhuwan and Liu, Zhengping and Cohen, William W and Lu, Xinghua},
  journal={arXiv preprint arXiv:1909.06146},
  year={2019}
}

@article{zhang2017chinese,
  title={Chinese medical question answer matching using end-to-end character-level multi-scale CNNs},
  author={Zhang, Sheng and Zhang, Xin and Wang, Hui and Cheng, Jiajun and Li, Pei and Ding, Zhaoyun},
  journal={Applied Sciences},
  volume={7},
  number={8},
  pages={767},
  year={2017},
  publisher={MDPI}
}

@article{he2019applying,
  title={Applying deep matching networks to Chinese medical question answering: A study and a dataset},
  author={He, Yun and Zhu, Ziwei and Zhang, Yin and Chen, Qin and Caverlee, James},
  journal={BMC Medical Informatics and Decision Making},
  volume={19},
  number={2},
  pages={91--100},
  year={2019},
  publisher={BioMed Central}
}

@article{zhang2022cblue,
  title={CBLUE: A Chinese biomedical language understanding evaluation benchmark},
  author={Zhang, Ningyu and Jia, Qianghuai and Yin, Kangping and Dong, Liang and Gao, Feng and Hua, Nengwei},
  journal={arXiv preprint arXiv:2106.08087},
  year={2022}
}

@article{brown2020language,
  title={Language models are few-shot learners},
  author={Brown, Tom and Mann, Benjamin and Ryder, Nick and Subbiah, Melanie and Kaplan, Jared D and Dhariwal, Prafulla and Neelakantan, Arvind and Shyam, Pranav and Sastry, Girish and Askell, Amanda and others},
  journal={Advances in Neural Information Processing Systems},
  volume={33},
  pages={1877--1901},
  year={2020}
}

@article{openai2023gpt4,
  title={GPT-4 technical report},
  author={OpenAI},
  journal={arXiv preprint arXiv:2303.08774},
  year={2023}
}

@article{singhal2022large,
  title={Large language models encode clinical knowledge},
  author={Singhal, Karan and Azizi, Shekoofeh and Tu, Tao and Mahdavi, S Sara and Wei, Jason and Chung, Hyung Won and Scales, Nathan and Tanwani, Ajay and Cole-Lewis, Heather and Pfohl, Stephen and others},
  journal={arXiv preprint arXiv:2212.13138},
  year={2022}
}

@article{singhal2023towards,
  title={Towards expert-level medical question answering with large language models},
  author={Singhal, Karan and Tu, Tao and Gottweis, Juraj and Sayres, Rory and Wulczyn, Ellery and Hou, Le and Clark, Kevin and Pfohl, Stephen and Cole-Lewis, Heather and Neal, Darlene and others},
  journal={arXiv preprint arXiv:2305.09617},
  year={2023}
}

@article{wu2023pmc,
  title={PMC-LLaMA: Towards building open-source language models for medicine},
  author={Wu, Chaoyi and Lei, Xiaoman and Chen, Yulei and Zhang, Xinyu and Yang, Jianfeng and Li, Kevin and Ma, Ke and Zheng, Shizhong and Xu, Xuyao and Zhou, Shuang K and others},
  journal={arXiv preprint arXiv:2304.14454},
  year={2023}
}

@article{li2023chatdoctor,
  title={ChatDoctor: A medical chat model fine-tuned on LLaMA model using medical domain knowledge},
  author={Li, Yunxiang and Li, Zihan and Zhang, Kai and Dan, Ruilong and Jiang, Steve and Zhang, You},
  journal={Cureus},
  volume={15},
  number={6},
  year={2023}
}

@article{xiong2023doctorglm,
  title={DoctorGLM: Fine-tuning your Chinese doctor is not a herculean task},
  author={Xiong, Honglin and Wang, Sheng and Zhu, Yitao and Zhao, Zihao and Liu, Yuxiao and Wang, Qian and Shen, Dinggang},
  journal={arXiv preprint arXiv:2304.01097},
  year={2023}
}

@article{labrak2024biomistral,
  title={BioMistral: A collection of open-source pretrained large language models for medical domains},
  author={Labrak, Yanis and Bazoge, Adrien and Morin, Emmanuel and Gourraud, Pierre-Antoine and Rouvier, Micka{\"e}l and Dufour, Richard},
  journal={arXiv preprint arXiv:2402.10373},
  year={2024}
}

@article{toma2023clinical,
  title={Clinical camel: An open-source expert-level medical language model with dialogue-based knowledge encoding},
  author={Toma, Augustin and Lawler, Patrick R and Ba, Jimmy and Krishnan, Rahul G and Rubin, Barry B and Volkovs, Maksims},
  journal={arXiv preprint arXiv:2305.12031},
  year={2023}
}

@article{zhang2023huatuogpt,
  title={HuatuoGPT, towards taming language model to be a doctor},
  author={Zhang, Hongbo and Chen, Junying and Jiang, Feng and Yu, Fei and Chen, Zhihong and Li, Jianquan and Chen, Guiming and Wu, Xiangbo and Zhang, Zhiyi and Xiao, Qingying and others},
  journal={arXiv preprint arXiv:2305.15075},
  year={2023}
}

@article{wang2023huatuo,
  title={Huatuo: Tuning LLaMA model with Chinese medical knowledge},
  author={Wang, Haochun and Liu, Chi and Xi, Nuwa and Qiang, Zewen and Zhao, Sendong and Qin, Bing and Liu, Ting},
  journal={arXiv preprint arXiv:2304.06975},
  year={2023}
}

@article{yang2024zhongjing,
  title={Zhongjing: Enhancing Chinese medical capabilities of large language model through expert feedback and real-world multi-turn dialogue},
  author={Yang, Songhua and Zhao, Hanjie and Zhu, Senbin and Zhou, Guangyu and Xu, Hongfei and Jia, Yuxiang and Zan, Hongying},
  journal={arXiv preprint arXiv:2308.03549},
  year={2024}
}

@misc{MedicalGPT2023,
  author = {Junying Chen and Xidong Wang and Ke Ji and Anningzhe Gao and Benyou Wang},
  title = {MedicalGPT: Training Medical GPT Model},
  year = {2023},
  publisher = {GitHub},
  journal = {GitHub repository},
  howpublished = {\url{https://github.com/shibing624/MedicalGPT}}
}

@article{fleming2024medalign,
  title={MedAlign: A clinician-generated dataset for instruction following with electronic medical records},
  author={Fleming, Scott L and Lozano, Alejandro and Haberkorn, William James and Jindal, Jenelle A and Reis, Eduardo Pontes and Thapa, Rahul and Blankemeier, Louis and Genkins, Julian Z and Steinberg, Ethan and Nayak, Ashwin and others},
  journal={arXiv preprint arXiv:2308.14089},
  year={2024}
}

@article{thirunavukarasu2023large,
  title={Large language models in medicine},
  author={Thirunavukarasu, Arun James and Ting, Darren Shu Jeng and Elangovan, Kabilan and Gutierrez, Laura and Tan, Ting Fang and Ting, Daniel Shu Wei},
  journal={Nature Medicine},
  volume={29},
  number={8},
  pages={1930--1940},
  year={2023},
  publisher={Nature Publishing Group}
}

@article{zhu2023multilingual,
  title={Multilingual large language models for biomedical question answering},
  author={Zhu, Mingchen and Ahuja, Aman and Wei, Wei and Reddy, Chandan K},
  journal={arXiv preprint arXiv:2305.13052},
  year={2023}
}

@article{wang2023cross,
  title={Cross-lingual transfer of medical language models},
  author={Wang, Xuezhi and Wei, Jason and Schuurmans, Dale and Le, Quoc and Chi, Ed and Narang, Sharan and Chowdhery, Aakanksha and Zhou, Denny},
  journal={arXiv preprint arXiv:2305.15329},
  year={2023}
}

@article{jiang2024mixtral,
  title={Mixtral of experts},
  author={Jiang, Albert Q and Sablayrolles, Alexandre and Roux, Antoine and Mensch, Arthur and Savary, Blanche and Bamford, Chris and Chaplot, Devendra Singh and Casas, Diego de las and Hanna, Emma Bou and Bressand, Florian and others},
  journal={arXiv preprint arXiv:2401.04088},
  year={2024}
}

@article{dai2024deepseekmoe,
  title={DeepSeekMoE: Towards ultimate expert specialization in mixture-of-experts language models},
  author={Dai, Damai and Deng, Chengqi and Zhao, Chenggang and Xu, R.X. and Gao, Huazuo and Chen, Deli and Li, Jiashi and Zeng, Wenge and Yu, Xingkai and Wu, Y. and others},
  journal={arXiv preprint arXiv:2401.06066},
  year={2024}
}

@article{gu2021domain,
  title={Domain-specific language model pretraining for biomedical natural language processing},
  author={Gu, Yu and Tinn, Robert and Cheng, Hao and Lucas, Michael and Usuyama, Naoto and Liu, Xiaodong and Naumann, Tristan and Gao, Jianfeng and Poon, Hoifung},
  journal={ACM Transactions on Computing for Healthcare},
  volume={3},
  number={1},
  pages={1--23},
  year={2021},
  publisher={ACM}
}

@book{bloom1956taxonomy,
  title={Taxonomy of educational objectives: The classification of educational goals},
  author={Bloom, Benjamin Samuel and others},
  year={1956},
  publisher={David McKay Company}
}

@misc{ollama2024,
  title={Ollama: Run large language models locally},
  author={Ollama},
  year={2024},
  howpublished={\url{https://ollama.ai}},
  note={Accessed: 2024-12-01}
}

\end{document}